\documentclass[letterpaper, 10 pt, conference]{ieeeconf} 
\IEEEoverridecommandlockouts
\input{config}

\title{\LARGE \bf
Floor Plan-Guided Visual Navigation Incorporating Depth and Directional Cues
}
\author{
 Weiqi Huang$^{1}$,
 Jiaxin Li$^{1}$,
 Zan Wang$^{1}$,
 Huijun Di$^{1}$,
 Wei Liang$^{1,2,\dag}$,
 Zhu Yang$^{1}$
 \thanks{$^{1}$Beijing Institute of Technology, $^{2}$Yangtze Delta Region Academy of Beijing Institute of Technology, $^{\dag}$indicates the corresponding author.}
}
\begin{document}
\maketitle
\begin{abstract}
Current visual navigation strategies mainly follow an exploration-first and then goal-directed navigation paradigm. This exploratory phase inevitably compromises the overall efficiency of navigation. Recent studies propose leveraging floor plans alongside RGB inputs to guide agents, aiming for rapid navigation without prior exploration or mapping. Key issues persist despite early successes. The modal gap and content misalignment between floor plans and RGB images necessitate an efficient approach to extract the most salient and complementary features from both for reliable navigation. Here, we propose \model, a novel framework that employs a diffusion-based policy to continuously predict future waypoints. This policy is conditioned on two complementary information streams: (1) local depth cues derived from the current RGB observation, and (2) global directional guidance extracted from the floor plan. The former handles immediate navigation safety by capturing surrounding geometry, while the latter ensures goal-directed efficiency by offering definitive directional cues. Extensive evaluations on the FloNa benchmark demonstrate that \model achieves superior efficiency and effectiveness. Furthermore, its successful deployment in real-world scenarios underscores its strong potential for broad practical application.
\end{abstract}

\section{Introduction}
Indoor visual navigation has long been a prominent topic in embodied AI research, fostering a wide range of downstream applications, including emergency rescue, assistive robotics, and autonomous exploration. At its core, the task requires agents to navigate through diverse, previously unseen environments toward specified goals using visual observations.
These observations, often in the form of RGB or RGB-D images, provide only limited egocentric views of the environment, thereby constraining the agent's ability to plan globally optimal and efficient paths.

\begin{figure}[t]
    \centering
    \includegraphics[width=\linewidth]{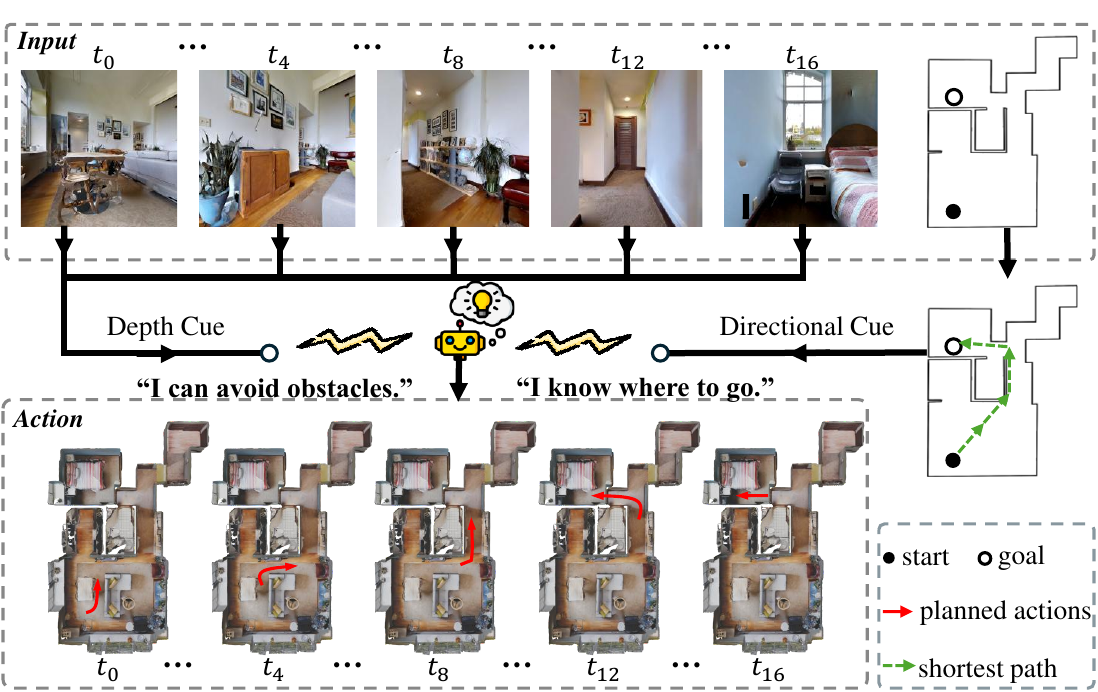}
    \caption{At each time step, our method extracts the depth cue from the observation frames and derives the directional cue by planning the shortest path to the goal on the floor plan. Both complementary cues facilitate the learning of a navigation policy that integrates obstacle avoidance with efficient goal-reaching behavior.}
    \label{fig:teaser}
\end{figure}

To address this limitation, recent work \cite{li2021cognitive, ewe2024spatial, li2024flona, setalaphruk2003robot, ericson2024beyond} has increasingly explored integrating prior floor plans with visual inputs. Floor plans constitute a valuable and widely available source of global spatial knowledge, providing high-level semantic and geometric layouts that remain invariant to dynamic changes such as temporary obstacles or furniture rearrangement. This inherent stability makes them well-suited as priors for supporting navigation in unseen environments.

However, effectively integrating these two data sources for both local obstacle avoidance and global path planning remains challenging. The main bottleneck is the inherent mismatch between them: (1) a difference in the way spatial information is represented in egocentric observations versus abstract top-down architectural plans, and (2) a content gap, where objects (e.g., furniture) present in observations are missing from the static floor plan. While prior approaches bypass this challenge by establishing extra topological or semantic maps, combining them with perceptions from LiDAR \cite{ewe2024spatial, li2021cognitive} and inertial measurement units \cite{goswami2024floor}, they also introduce increased system costs. Li et al. \cite{li2024flona} employ a unified image encoder to extract latent features from both floor plans and visual observations, followed by transformer-based fusion for action prediction. This method achieves only a superficial and ineffective fusion. By processing the floor plan as a mere image, it squanders the inherent, explicit representation of global connectivity that is the defining feature of a human-abstracted topological map.

We propose \model, a diffusion-based policy for floor plan-guided visual navigation. The core idea of \model lies in leveraging depth and directional cues from RGB observations and the floor plan, respectively, to guide action generation. As shown in \cref{fig:teaser}, \model bridges the representational gap by extracting a sequence of 2D waypoints (the shortest path) from the floor plan's topology. This path—despite potential obstacle intersections—provides a geometrically grounded directional prior, translating the abstract map into actionable sub-goals for egocentric control. Concurrently, \model complements the global path with fine-grained local geometry. It leverages a depth encoder \cite{ke2024repurposing} to produce depth features from RGB observations, which reveal traversable regions and obstacles absent from the floor plan. To synthesize smooth and temporally coherent action sequences from these complementary signals, we employ a conditional diffusion model. It is conditioned on both guidance sources and trained to generate actions by supervised learning on optimal trajectories. To enhance robustness against the inevitable localization errors known to impact navigation performance \cite{li2021cognitive, li2024flona}, we incorporate Gaussian noise perturbations into the agent’s pose and path planning results during training. Consequently, \model can be effectively deployed with a standard VO module \cite{freda2025pySLAM}, achieving robust performance even when faced with its noisy pose estimates.
Extensive experiments on the FloNa \cite{li2024flona} benchmark demonstrate that our approach outperforms the baselines even with fewer training pairs, emphasizing its effectiveness and data efficiency. We also conduct comprehensive ablation studies to assess the impact of conditioning on the global shortest path and local depth features, as well as \model's robustness to varying localization accuracy. Finally, we validate the practical applicability of \model by deploying it on an Automated Guided Vehicle (AGV) without any fine-tuning, showcasing its potential for real-world deployment.

Our contributions are summarized as follows:
\begin{itemize}
\item We propose conditioning the navigation policy on global path planning and local depth features to provide more effective guidance, thus enabling efficient and collision-free navigation.
\item Extensive experiments validate the effectiveness and generalization of our model. Additionally, the noise injection strategy during training ensures robust adaptation to VO, leading to successful deployment on the real robot.
\end{itemize}

\begin{figure*}[t!]
    \centering
    \includegraphics[width=\textwidth]{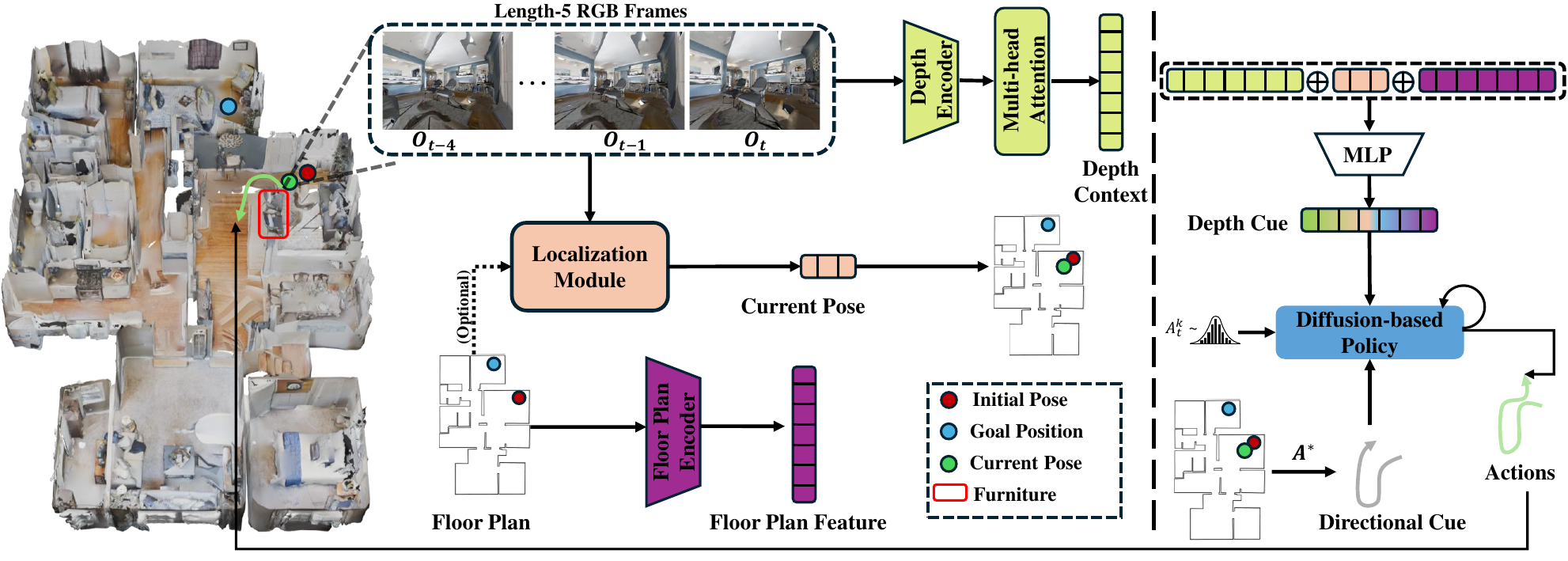}
    \caption{\textbf{Overview of \model.} The pre-trained depth encoder uses RGB observations to produce depth latent codes, which are fed into a multi-head attention module, yielding the depth context. Taking depth context, current pose from the localization module, and floor plan feature from the floor plan encoder as inputs, the MLP outputs geometry guidance. Path planner computes the shortest path from the current position to the goal using the A$^*$ algorithm, serving as the path guidance. Conditioned on geometry and path guidance, the diffusion-based policy generates future actions.
    }
    \label{fig:pipeline}
\end{figure*}

\section{Related Work}
\subsection{Mapping-based Navigation}
 Mapping-based navigation, the prevailing framework, requires building an environmental map first to support subsequent planning and localization. Typically, VSLAM techniques \cite{mur2015orb, mur2017orb} rely on visual observation inputs to construct sparse metric maps of the surrounding environment. In recent years, more works have started to integrate semantic information into map representations. Researchers \cite{cha2020obj, ram2022pon, guo2024object} extract semantic categories from observations, associate the categories with each point in the point cloud, and finally generate a semantic map via differentiable projection operations and a denoising network. 
Other approaches integrate point-level features \cite{huang2023visual}, image features \cite{yoo2024common, chen2022think, wang2023dreamwalker}, and panoramic images \cite{chaplot2020neural} into the built topological graph. VLFM \cite{yokoyama2024vlfm} iteratively constructs an occupancy map and a value map to guide the agent toward the out-of-view target object. However, the mapping-based method has to explore unseen areas, which tends to be time-consuming for navigation. Moreover, the dynamic nature of objects in indoor environments necessitates real-time updates to the constructed map, making its reuse difficult and significantly reducing efficiency.

\subsection{Mapless Navigation}
Without relying on mapping, mapless visual navigation methods directly focus on translating observations into actions. Wjimans et al. \cite{wijmans2020ddppo} solve the point goal navigation with a near-perfect success rate under idealized settings through large-scale reinforcement learning. Learning-based visual odometry \cite{partsey2022ismapping, zhao2021surprising} is used to address realistic point goal navigation. Wenzel et al. \cite{wenzel2021vision} train visual-motor policies for navigation tasks in a simple 3D environment. Work \cite{ran2021scene} design a lightweight perception network and convert visual navigation to scene classification. Leveraging the diffusion policy \cite{chi2023diffusion}, NoMaD \cite{sridhar2024nomad} produces collision-free future actions and flexibly handle both goal-conditioned and goal-agnostic navigation via the goal masking technique. Although mapless methods eliminate the need of mapping, they often produce shortsighted actions due to the lack of global navigable information.

\subsection{Navigation with Floor Plan}
Unlike the two methods mentioned above, floorplan-based navigation eliminates the need to explore the environment and instead utilizes the inherent structure and spatial information to perform reliable navigation actions. Setalaphruk et al. \cite{setalaphruk2003robot} extract a Voronoi diagram from the sketch map and subsequently implement localization and navigation based on multiple hypotheses.
Li et al. \cite{li2021cognitive} propose a cognitive navigation solution using floor plans. However, its dependency on pre-collect visual features hinders its practical applicability. To eliminate the dependency, Ewe et al. \cite{ewe2024spatial} develop a graph-based navigation framework, enabling navigation without relying on precise scale information. Goswami et al. \cite{goswami2024floor} leverage stereo-inertial sensors to construct 2D semantic point cloud, which, in conjunction with the floor plan, enables robot localization and subsequent navigation.
These methods all utilize sensors other than monocular cameras. The closest related work to ours is FloNa \cite{li2024flona}, which marks the first attempt to navigate unseen environments using only a monocular camera and a floor plan. FloNa utilizes a transformer-based vision encoder to extract observation context from the floor plan and visual inputs. A diffusion policy \cite{chi2023diffusion}, conditioned on the context, is then employed to generate actions.
Although FloNa demonstrates impressive performance, it fails to effectively integrate complementary geometric clues from observations and floor plans. \model tackles this issue by independently processing global and local information to derive two distinct guidances.

\subsection{Diffusion Policy} 
Diffusion-based policies have been successfully applied in various robotic domains, including dexterous manipulation \cite{simeonov2023rpdiff, chi2023diffusion}, locomotion \cite{huang2024diffuseloco}, and path planning \cite{sridhar2024nomad, li2024flona}.
By framing robot behavior generation as a conditional denoising diffusion process, these policies not only offer high training stability but also demonstrate strong performance in modeling complex distributions. Building upon these advances, our work focuses on utilizing diffusion policies conditioned on geometric clues derived from both visual observations and floor plans to achieve vision-based floor plan navigation.

\section{Methodology}
As illustrated in \cref{fig:pipeline}, \model extracts a global directional cue from the floor plan and a local depth cue primarily from the RGB observation, which together guide the diffusion policy for future action generation. In the following, we detail each key module of the proposed framework.
\subsection{Depth Cue Module}
To guide \model in generating collision-free actions, the depth cue module processes RGB frames and produces low-dimensional local geometric guidance.

Specifically, we employ a depth encoder \cite{ke2024repurposing} to extract depth features from RGB images. This encoder utilizes a latent diffusion model to model the conditional distribution $P(\mathbf{d}|\mathbf{o})$, where $\mathbf{d}$ and $\mathbf{o}$ denote the low-dimensional depth feature and RGB observation, respectively. In our model, the depth encoder takes $5$ observations as input, including the current observation and the past $4$ observations, denoted as $\mathcal{O}=\{o_{t-4}, o_{t-3},\ldots, o_{t}\}$. Then, the encoder outputs corresponding depth features denoted as $\mathcal{D}=\{d_{t-4},d_{t-3},\ldots,d_{t}\}$. These features are then passed through the multi-head attention module \cite{vaswani2017attention}, and the resulting fused features are averaged to derive the depth context. Subsequently, following \cite{li2024flona}, we employ a floor plan encoder consisting of an EfficientNet-B0 \cite{tan2019efficientnet} network and a multilayer perceptron (MLP) to encode the floor plan image,  obtaining a floor plan feature. Finally, the depth context, the floor plan feature, and the agent's pose on the floor plan from the localization module (see \cref{localization module}) are then concatenated and fed into an MLP to produce the local depth cue, $C_l$.

\subsection{Localization Module}
\label{localization module}
In our work, we employ a localization module to determine the current pose of the agent on the floor plan, $p_{f}$, using the current ego-centric observations. $p_{f}$ is represented as $p_{f} = (q_{f}, r)$, where $q_{f}=(x_{f}, y_{f})$ denotes the pixel coordinate on the floor plan, and $r$ represents the orientation. In this task, the floor plan is scaled to the environment, enabling the following equation to relate world coordinates to pixel coordinates:
\begin{equation}
    p_{w} = (q_{w}, r) = (\mu \times x_{f} + \delta_x, \mu \times y_{f} + \delta_y, r),
    \label{eq: world2pixel}
\end{equation}
where $\mu$ denotes the floor plan resolution (\ie, physical distance per pixel) and $\delta$ represents the offset.

During training, we use the ground truth (GT) poses as the localization results. For testing and deployment, we utilize a visual odometry or a floor plan-based visual localization module to estimate the current pose.

\subsection{Directional Cue Module}
The directional cue module provides the policy with global shortest path guidance. 
Specifically, we employ the A$^*$ algorithm to compute the shortest path from the current position to the target within the floor plan. The path is represented as $\mathcal{T}_s=\{q_{f,0}, q_{f,1}, \cdots, q_{f,i}, \cdots\, q_{f,t}\}$, where $q_{f,0}$ denotes the pixel coordinates of the current position and $q_{f,t}$ corresponds to the target point.
Then, we convert each point in $\mathcal{T}_s$ into metric coordinate using $\mu$ and $\delta$ in \cref{eq: world2pixel}, represented as $\mathcal{T}_{w,s}=\{q_{w,0}, q_{w,1}, \cdots, q_{w,i}, \cdots\, q_{w,t}\}$.
Next, we transform each point in $\mathcal{T}_{w,s}$ to the current agent coordinate system by utilizing the formula
\begin{equation}
    q_{a,i}= (q_{w,i} - q_{w,0}) \cdot R,
    \label{eq: world2agent}
\end{equation}
resulting in $\mathcal{T}_{a,s} = \{q_{a,0}, q_{a,1}, \cdots, q_{a,i}, \cdots, q_{a,t}\}$.
Here, $R = \bigl( \begin{smallmatrix} \cos{r_0} & -\sin{r_0}\\ \sin{r_0} & \cos{r_0}\end{smallmatrix} \bigr)$, $r_0$ denotes the orientation of the first point in the shortest path. In the agent coordinate system, the agent's current position serves as the coordinate origin, the current orientation aligns with the positive X-axis, while the leftward direction aligns with the positive Y-axis. Notably, we assume the agent does not experience height variation during navigation, and thus, we only consider a 2D coordinate system (without the Z-axis). 
Finally, we take the first $m$ points of $\mathcal{T}_{a,s}$ (about $1.2$ m) as the path guidance $C_{g}$.
\subsection{Diffusion Policy}
Diffusion policy \cite{chi2023diffusion} leverages the Denoising
Diffusion Probabilistic Models (DDPMs) \cite{ho2020denoising} paradigm to formulate visuomotor robot policies. In our work, using the depth and directional cues, $C_l$ and $C_g$, the diffusion policy learns to approximate the conditional distribution $P(A_t|C_l, C_g)$ through a forward and reverse process, where $A_t$ is a $m\times2$ vector representing the generated sequence of future $m$ actions at time step $t$. Each action corresponds to an approximate movement of $4$ cm.

In the forward process, diffusion policy iteratively adds Gaussian noise and converts the $A_t^0$ to $A_t^K$, where $A_t^0$ is the initial action sequence, and $K$ is the total steps. The entire process is given by $P(A^{1:K}_t|A_t^0)=\prod_{k=1}^{K}P(A_t^k|A_t^{k-1})$.

\begin{figure}[t!]
    \centering
    \includegraphics[width=\linewidth]{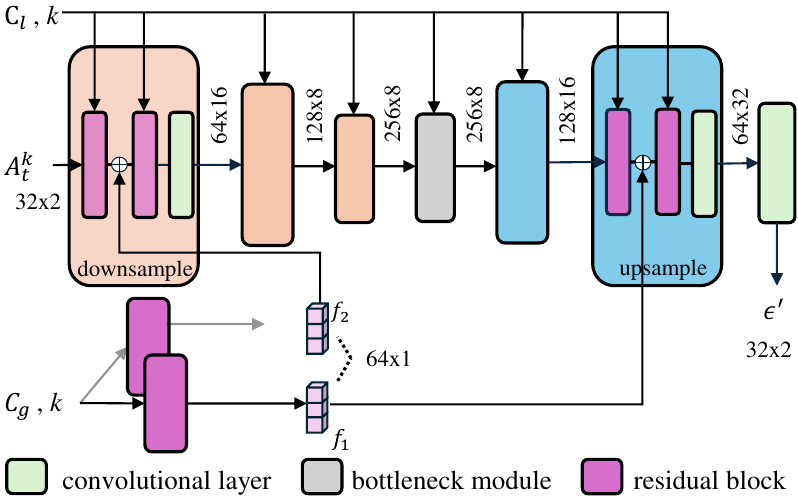}
    \caption{\textbf{Conditional UNet.} The directional cue and the time step are fed into the first down module and the second up module, while the depth cue and the time step are fed into each module.}
    \label{fig: conditional unet}
\end{figure}

In the reverse process, diffusion policy learns the conditional distribution $P(A_t^{k-1}|A_t^k, C_l, C_g)$. Specifically, the diffusion policy employs a noise prediction network of UNet \cite{ronneberger2015unet} architecture to predict the added noise, $\epsilon^\prime$, and utilizes the noise scheduler to carry out the denoising operation. The process can be described as:
\begin{equation}
    \epsilon^\prime = \varepsilon_{\theta}(A_t^k, C_l, C_g,k),
\end{equation}
\begin{equation}
    A_t^{k-1}=\mathcal{S}(\epsilon^\prime, A_t^k,k),
\end{equation}
where $\varepsilon$ denotes the noise prediction network with parameters $\theta$ and $\mathcal{S}$ is the DDPM noise scheduler.
As shown in \cref{fig: conditional unet}, our noise prediction network includes three downsample modules, one bottleneck module, two upsample modules, and a 1D convolutional layer. Each module consists of several 1D convolutional blocks, activation functions, and residual connections. The network takes noisy actions of size $m\times2$, along with depth and path guidance, $C_l$ and $C_g$,  as input and outputs a predicted noise of size $m\times2$. $C_l$ and the time step serve as inputs to each module. Meanwhile, $C_g$ and the time step are first processed through two identical residual blocks, producing two features: $f_1$ and $f_2$. $f_1$ is added to the output of the first residual block in the final upsample module, while $f_2$ is added to the output of the first residual block. Notably, we apply positional encoding to the time step before feeding it into the network.
During training, we use the following loss function:
\begin{equation}
    \mathcal{L} = \texttt{MSE}(\epsilon, \epsilon^\prime),
\end{equation}
where $\texttt{MSE}(\cdot)$ denotes the mean square error function, and $\epsilon$ denotes the added noise at the forward process. During inference, our model iteratively predicts and removes noise to recover a clean action sequence $A_t^0$ from Gaussian noise.

\begin{table*}[t!]
    \centering
    \caption{SR ($\%$) and SPL ($\%$) of Different Models under Various Conditions. The \textbf{Bold} Number Indicates the Best Result.}
    \label{tab:results}
    \resizebox{\linewidth}{!}{%
    \begin{tabular}{ccccccccccccccccccc}%
        \toprule 
        \multicolumn{3}{c}{\multirow{2}[4]{*}{Method}} & \multicolumn{2}{c}{Loc-A* (GT)}& \multicolumn{2}{c}{DD-PPO (GT)} & \multicolumn{2}{c}{Map-A* (GT)} & \multicolumn{2}{c}{Loc-FloDiff (F$^3$)} & \multicolumn{2}{c}{Loc-FloDiff (GT)} & \multicolumn{2}{c}{\model (F$^3$)}& \multicolumn{2}{c}{\model (VO)} & \multicolumn{2}{c}{\model (GT)} \\
        \cmidrule(r){4-19}
        & & & Gibson & HM3D & Gibson & HM3D & Gibson & HM3D & Gibson & HM3D & Gibson & HM3D & Gibson & HM3D & Gibson & HM3D & Gibson & HM3D \\
        \midrule
        \multicolumn{1}{c|}{\multirow{4}{*}{SR(\%) $\uparrow$}} & \multicolumn{1}{c}{\multirow{4}{*}{$\tau_c$}}
                                                      & $10$    & $16.60$ & $18.60$ & $15.00$  & $13.20$ & $55.60$  & $46.80$  & $4.80$   & $2.40$ & $39.00$  & $34.40$  & $53.40$ & $47.60$ & $68.60$  & $63.40$     & \bm{$79.60$}  & \bm{$77.60$}\\
        \multicolumn{1}{c|}{} & \multicolumn{1}{c}{}  & $30$    & $18.20$ & $18.80$ & $22.80$  & $16.40$ & $55.60$  & $46.80$  & $11.60$  & $5.20$ & $53.40$  & $45.80$  & $69.20$ & $55.80$ & $80.80$  & $71.80$     & \bm{$92.80$}  & \bm{$90.40$}\\
        \multicolumn{1}{c|}{} & \multicolumn{1}{c}{}  & $50$    & $18.50$ & $19.40$ & $31.60$  & $23.80$ & $55.60$  & $46.80$  & $17.80$  & $9.20$ & $59.20$  & $48.40$  & $77.80$ & $62.60$ & $86.60$  & $79.00$     & \bm{$94.20$}  & \bm{$90.80$}\\
        \multicolumn{1}{c|}{} & \multicolumn{1}{c}{}  &$\infty$ & $18.60$ & $19.40$ & $48.20$  & $36.00$ & $55.60$  & $46.80$  & $26.80$  & $15.80$& $66.00$  & $53.20$  & $86.40$ & $68.40$ & $89.80$  & $83.60$     & \bm{$97.60$}  & \bm{$91.80$}\\             
        \midrule
        \multicolumn{1}{c|}{\multirow{4}{*}{SPL(\%)$\uparrow$}} & \multicolumn{1}{c}{\multirow{4}{*}{$\tau_c$}} 
                                                      & $10$    & $16.50$  & $17.44$ & $8.40$  & $7.80$  & $34.23$  & $33.79$  & $4.02$   & $1.54$ & $36.30$  & $30.60$  & $41.67$ & $37.85$ & $52.88$  & $46.84$     & \bm{$62.62$}  & $61.24$\\
        \multicolumn{1}{c|}{} & \multicolumn{1}{c}{}  & $30$    & $18.10$  & $17.80$ & $12.60$ & $9.60$  & $34.23$  & $33.79$  & $6.40$   & $3.36$ & $44.10$  & $35.28$  & $49.35$ & $41.11$ & $59.07$  & $52.43$     & \bm{$70.38$}  & \bm{$69.88$}\\
        \multicolumn{1}{c|}{} & \multicolumn{1}{c}{}  & $50$    & $18.50$  & $18.35$ & $16.40$ & $13.80$ & $34.23$  & $33.79$  & $8.07$   & $5.18$ & $46.60$  & $37.20$  & $52.61$ & $47.55$ & $61.31$  & $55.04$     & \bm{$70.98$}  & \bm{$70.31$}\\
        \multicolumn{1}{c|}{} & \multicolumn{1}{c}{}  &$\infty$ & $18.50$  & $18.35$ & $23.00$ & $18.20$ & $34.23$  & $33.79$  & $9.58$   & $8.44$ & $48.20$  & $40.41$  & $54.77$ & $49.75$ & $62.57$  & $56.49$     & \bm{$72.36$}  & \bm{$71.69$}\\
        \bottomrule
    \end{tabular}%
    }%
\end{table*}

\subsection{Training}
\label{training}
\label{training}
We use the data provided by FloNa \cite{li2024flona}, collected from $67$ Gibson simulation environments. Each scene contains $150$, $180$, or $200$ episodes, depending on the scene size, resulting in a total of $20,214$ episodes. We sample training data from this dataset, with each sample consisting of a historical observation of five consecutive RGB frames, the agent’s current position, a goal position, a floor plan image, and the corresponding sequence of future waypoints. Unlike FloNa \cite{li2024flona}, which densely samples goal positions from future waypoints along the trajectory, we define the goal as the final waypoint of the episode. We sample one pair every five frames to avoid overlapping observations across different pairs. Consequently, our method is trained on 1.2 million samples, significantly fewer than the 180 million samples used for training Loc-FloDiff.

To improve the model's robustness to localization noise, we introduce perturbations to both the shortest path $\mathcal{T}_s$ and the ground truth current pose $p_{w}$ during training. Specifically, we perturb the shortest path using the following formula:
\begin{equation}
    \mathcal{T}_n = \mathcal{T}_s + u,
\end{equation}
where $\mathcal{T}_n$ denotes noisy path and  $u\sim\mathcal{U}(0,B)$ represents the perturbation. $B$ represents the upper bound of the perturbation, which increases with each epoch $i$, as shown by the following formula:
\begin{equation}
    \mathcal{B}_i = \alpha \cdot sin(\frac{i}{N}\cdot\frac{\pi}{2}),
\end{equation}
where $\alpha$ is the maximum perturbation value, while $N$ represents the total epochs. Meanwhile, we add Gaussian noise, $n_p\sim\mathcal{N}(0,0.1)$ and $n_r\sim\mathcal{N}(0,\frac{\pi}{36})$, to the ground truth position and orientation.

\subsection{Implementation}
In our implementation, the multi-head attention consists of four heads, each with four layers. We choose Square Cosine Noise Schedule \cite{nichol2021improved} to train the diffusion policy. We use the AdamW optimizer, with the learning rate controlled by PyTorch's CosineAnnealingLR scheduler. The maximum learning rate is set to $0.0001$, and the maximum number of iterations is $20$. We set $\alpha$ to $0.1$. The training of \model is performed on $8$ NVIDIA $4090$ GPUs, with a batch size of $64$ per GPU. \model is trained for $N=20$ epochs and reaches convergence.

\section{Experiments}
We evaluate our method through comprehensive simulation experiments (including baseline comparisons and ablation studies) and subsequently deploy the policy on a physical AGV.

\begin{table}[t!]
    \centering
    \caption{Baseline Methods and Their Components.}
    \label{tab:2}
    \resizebox{\linewidth}{!}{%
    \begin{tabular}{l|l|l|l}%
        \toprule
        \textbf{Model} & \textbf{Localization} & \textbf{Planner} & \textbf{Train Set Size} \\
        \midrule
        Loc-A* (GT)          & GT Pose                       & A*          & -      \\ 
        DD-PPO (GT)          & GT Pose                       & DD-PPO      & $2.5$B \\
        Map-A* (GT)          & GT Pose                       & A*          & -      \\
        Loc-FloDiff (GT)     & GT Pose                       & Loc-FloDiff & $180$M \\
        Loc-FloDiff (F$^3$)  & F$^3$Loc                      & Loc-FloDiff & $180$M \\
        GlocDiff (F$^3$)     & F$^3$Loc                      & GlocDiff    & $1.2$M \\
        GlocDiff (GT)        & GT Pose                       & GlocDiff    & $1.2$M \\
        GlocDiff (VO)        & Visual Odometry               & GlocDiff    & $1.2$M \\
        \bottomrule
    \end{tabular}%
    }%
\end{table}

\begin{table}[t]
    \vspace{-3pt}
    \centering
    \caption{Error Comparison of Localization Methods}
    \label{tab:f3&vo}
    \resizebox{\linewidth}{!}{%
        \begin{tabular}{ccccccc}%
            \toprule
            \multirow{2}{*}{\textbf{localization method}} & \multicolumn{3}{c}{\textbf{Translation (m)}} & \multicolumn{3}{c}{\textbf{Orientation ($^\circ$)}}\\
            \cmidrule(lr){2-4}\cmidrule(lr){5-7}
             & mean & median & max & mean & median & max\\
            \midrule
            \multicolumn{1}{c|}{\ac{vo}}       & $1.53$ & $1.66$ & $4.28$ & $13.35$ & $18.72$ & $22.51$\\            
            \multicolumn{1}{c|}{F$^3$Loc} & $3.44$ & $2.17$ & $9.47$ & $18.14$ & $17.96$ & $30.64$\\ 
            \bottomrule
        \end{tabular}%
}%
\end{table}

\subsection{Main Experiments}

\paragraph{Setup} 
We conduct our experiments in the Gibson \cite{xia2018gibson} simulator.
The testing scenes consist of 50 Gibson scenes and 50 HM3D \cite{ramakrishnan2021hm3d} scenes, each containing 10 episodes. The episodes for Gibson scenes are sourced from FloNa \cite{li2024flona}, while those for HM3D scenes are newly collected under the constraint that the Euclidean distance between the start and target points exceeds 10$m$ to ensure sufficient task difficulty.
The navigation is considered successful only if the agent arrives within a distance of $\tau_d$ from the target, within the maximum step limit, and with the total number of collisions below the threshold $\tau_c$. The agent responds to collisions by rotating 45 degrees clockwise and then re-predicting its future action sequence.
\paragraph{Baselines}
We set eight baselines, with their specific components outlined in \cref{tab:2}. \textbf{Loc-A* (GT):} The agent follows a path planned by the A* algorithm, with GT pose used for localization. \textbf{DD-PPO \cite{wijmans2020ddppo} (GT):} This baseline is a reinforcement learning–based agent that navigates using only visual observations, with GT pose for localization. \textbf{Map-A* (GT):} This baseline first uses the same depth estimation model \cite{ke2024repurposing} as \model to convert RGB observations into depth. The depth is then transformed into a point cloud, which is projected to form a bird's-eye-view (BEV) occupancy grid map. Navigation is carried out by planning a path with the A* algorithm over this occupancy map, using ground truth (GT) pose for localization.
Both \textbf{Loc-FloDiff (F$^3$)} and \textbf{GlocDiff (F$^3$)} employ F$^3$Loc \cite{chen2024f3loc}, an end-to-end floor plan localization model, as their localization module. \textbf{Loc-FloDiff (GT)} and \textbf{GlocDiff (GT)}, in contrast, rely on GT pose for localization. In \textbf{GlocDiff (VO)}, a VO algorithm \cite{freda2025pySLAM} is used to track the agent’s pose, serving as the localization module. For evaluation, we adopt Success Rate (SR) and Success Weighted by Path Length (SPL) as our metrics. We evaluate all baselines under the distance threshold $\tau_d$ values of $0.25$, as well as the collision threshold $\tau_c$ values of $10$, $30$, $50$, and $\infty$. 

\begin{figure*}[t!]
    \centering
    \includegraphics[width=\textwidth]{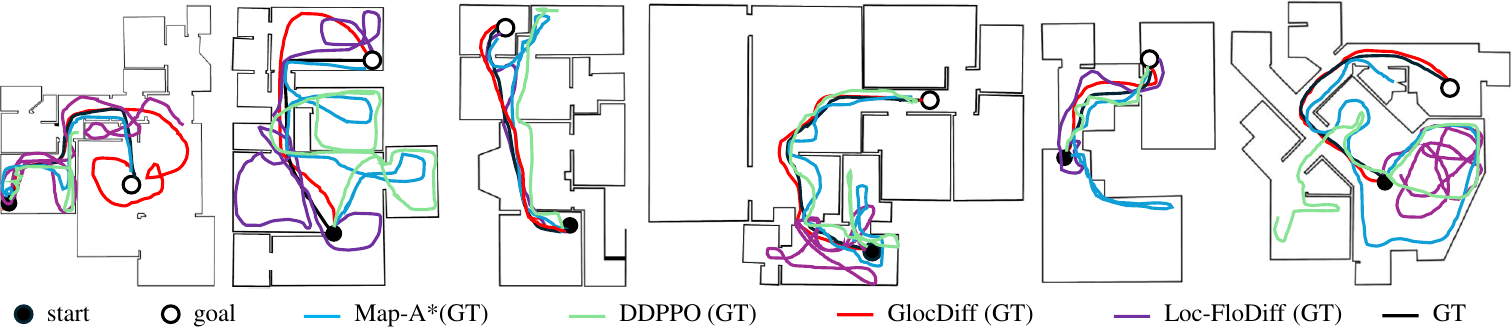}
    \caption{\textbf{Trajectory visualization.} The visualization of trajectories for different models across six episodes.}
    \label{fig:qualitative_results}
\end{figure*}

\begin{figure}[t!]
    \centering
    \includegraphics[width=\linewidth]{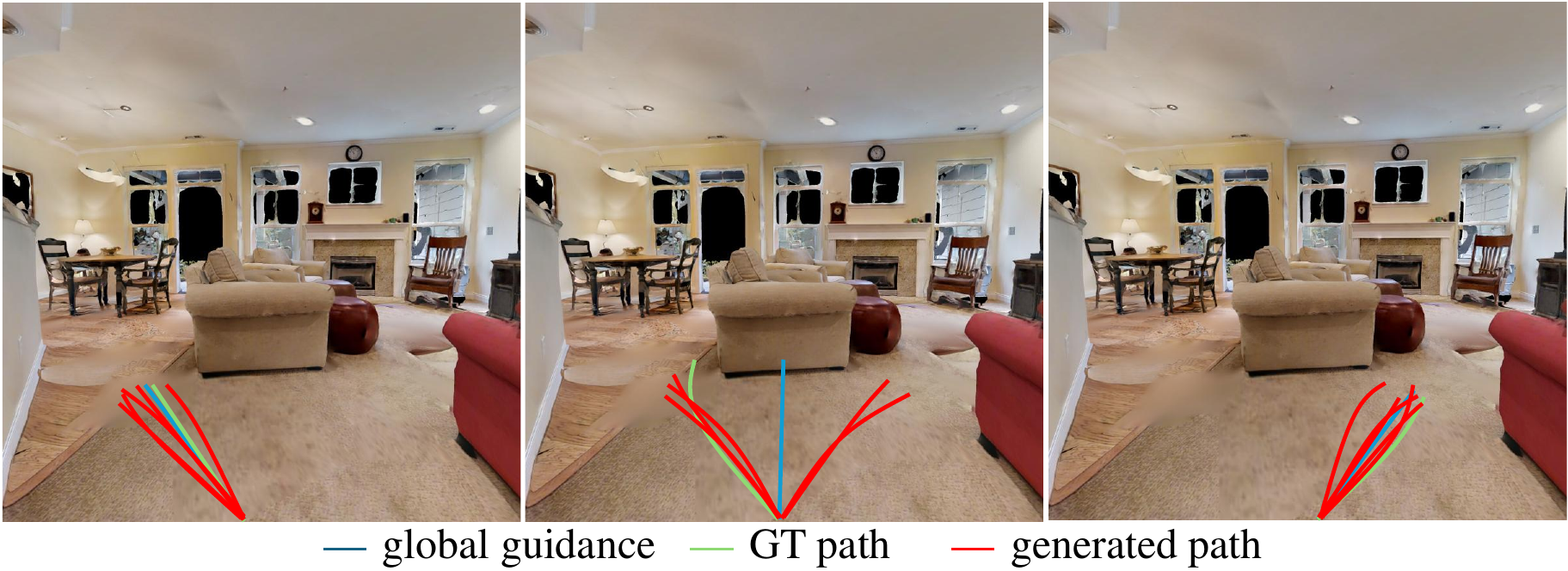}
    \caption{Trajectory visualization of \model (GT) under three distinct global guidance conditions for the same observation.}
    \label{fig:qualitative_2}
\end{figure}

\paragraph{Quantitative Results}
\cref{tab:results} presents the qualitative results. Below, we discuss several key observations:
\begin{itemize}[leftmargin=*,nolistsep,noitemsep]
\item DD-PPO (GT) and Map-A* (GT) consistently underperform \model (GT) across all settings. While Map-A* (GT) achieves slightly higher SR than Loc-FloDiff (GT) at $\tau_c = 10$ and $30$, its SPL is lower in all cases. This suggests that incorporating a floor plan significantly improves navigation efficiency. Notably, the poor performance of DD-PPO (GT) stems from its lack of explicit obstacle-avoidance capability, combined with our collision-sensitive evaluation—unlike the setup in \cite{wijmans2020ddppo}, which allows the agent to slide along obstacles.
\item Despite being trained on a significantly smaller dataset, \model consistently outperforms Loc-FloDiff by a notable margin in both success rate and SPL, regardless of whether localization is based on GT positions or F$^3$ Loc \cite{chen2024f3loc}. Moreover, when $\tau_c \geq 30$, \model (GT) attains a success rate above $90\%$ and an SPL exceeding about $70\%$, strongly indicating that our model effectively utilizes floor plan information and exhibits robust planning capability. These outcomes validate the effectiveness and high data efficiency of our approach. Additionally, when evaluated on HM3D scenes — which differ from the Gibson scenes used in training — \model exhibits a smaller performance decline than Loc-FloDiff. This indicates that extracting both shortest-path guidance and depth cues from floor plans and RGB observations generalizes more robustly across varied environments compared to methods that rely primarily on learned image features.
\item When the localization results shift from GT poses to predictions from the F$^3$Loc module, Loc-FloDiff experiences a significant decline in performance, with the SR dropping by $32.00\%$ to $41.8\%$ and the SPL decreasing by $29.06\%$ to $38.62\%$. In contrast, the performance of \model shows a more moderate decline, with the SR falling by $11.20\%$ to $34.60\%$ and the SPL decreasing by $17.59\%$ to $28.77\%$. This highlights that our model exhibits greater robustness to localization noise.  
\item The SR of \model (VO) is higher than that of \model (F$^3$), with the difference being particularly pronounced under stricter collision constraints (i.e., lower $\tau_c$).  This suggests that visual odometry (VO) serves as a more suitable localization module for floor plan navigation compared to F$^3$Loc. Since VO can continuously track the agent's pose, it remains relatively stable for indoor navigation tasks, despite cumulative errors. In contrast, F$^3$Loc maps observations to poses, introducing room-level errors, which is unacceptable for floor plan navigation. We further evaluate the localization accuracy of F$^3$ and VO. Specifically, we compute the mean and median errors between predicted and ground-truth waypoints across fifty sampled trajectories. As shown in \cref{tab:f3&vo}, VO achieves higher average accuracy than F$^3$Loc. Additionally, F$^3$Loc produces occasional outlier predictions, indicating abrupt jumps to distant locations in the environment.
\end{itemize}
\begin{table}[t!]
    \centering
    \caption{Ablation Study on Directional and Depth Cues.}
    \label{tab:ablation}
    \resizebox{\linewidth}{!}{%
        \begin{tabular}{cccccccc}%
            \toprule
            \multicolumn{2}{c}{\multirow{2}[4]{*}{Model}}& \multicolumn{2}{c}{w/o depth} & \multicolumn{2}{c}{w/o path} & \multicolumn{2}{c}{\model (GT)} \\
            \cmidrule{3-8}
            & & SR & SPL & SR & SPL & SR & SPL\\
            \midrule
            \multicolumn{1}{c}{\multirow{4}{*}{$\tau_c$}} & $10$      &  $64.20$  & $47.22$  & $44.60$  & $38.44$   & \bm{$79.60$}  & \bm{$62.62$} \\
            \multicolumn{1}{c}{}                          & $30$      &  $81.00$  & $55.98$  & $60.60$  & $44.09$   & \bm{$92.80$}  & \bm{$70.38$} \\
            \multicolumn{1}{c}{}                          & $50$      &  $85.20$  & $58.07$  & $67.00$  & $48.48$   & \bm{$94.20$}  & \bm{$70.98$} \\
            \multicolumn{1}{c}{}                          & $\infty$  &  $92.20$  & $60.15$  & $73.20$  & $50.51$   & \bm{$97.60$}  & \bm{$72.36$} \\
            \bottomrule
        \end{tabular}%
    }%
\end{table}
\paragraph{Qualitative Results}
\cref{fig:qualitative_results} presents qualitative trajectory visualizations of Map-A* (GT), DDPPO (GT), LocFlodiff (GT), and \model (GT) across six distinct scenarios. We further visualize five trajectories generated by \model (GT) under three different global guidance inputs for the same observation, as shown in \cref{fig:qualitative_2}. When the global guidance does not intersect with obstacles (left and right cases), \model (GT) consistently follows it closely across multiple runs, showing high confidence. In contrast, when the global guidance poses a collision risk with obstacles (middle case), \model (GT) places greater reliance on local guidance to navigate around them.

\subsection{Ablation Study}
\paragraph{Ablation Study on Local and Global Guidances}
We further evaluate the effects of different guidances on the policy. To assess the impact of the local guidance, we use an EfficientNet to directly encode the RGB observations, obtaining RGB context to replace the original depth context, while keeping other design aspects consistent with those of \model. The resulting ablation model is denoted as w/o depth. To analyze the influence of the global guidance, we eliminate the global planner and directly utilize the target point to replace the shortest path, while also keeping other aspects consistent with \model. This modification results in another ablation model, referred to as w/o path. The two models described above are both trained for $20$ epochs and reach convergence.

\begin{figure}[t!]
    \centering
    \includegraphics[width=\linewidth]{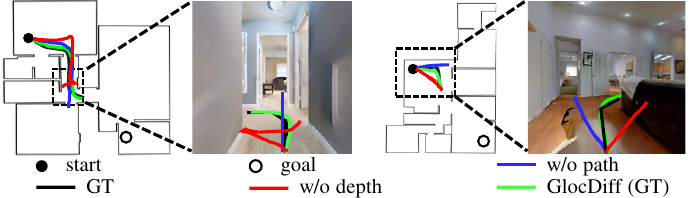}
    \caption{\textbf{Performance of different models on the test episodes in Mobridge (left) and Brentsville (right).} The left column illustrates the traversed trajectory, while the right column showcases the diverse actions generated by different models based on the same observation. }
    \label{fig:ablation}
\end{figure}

\begin{figure*}[t!]
    \centering
    \includegraphics[width=\linewidth]{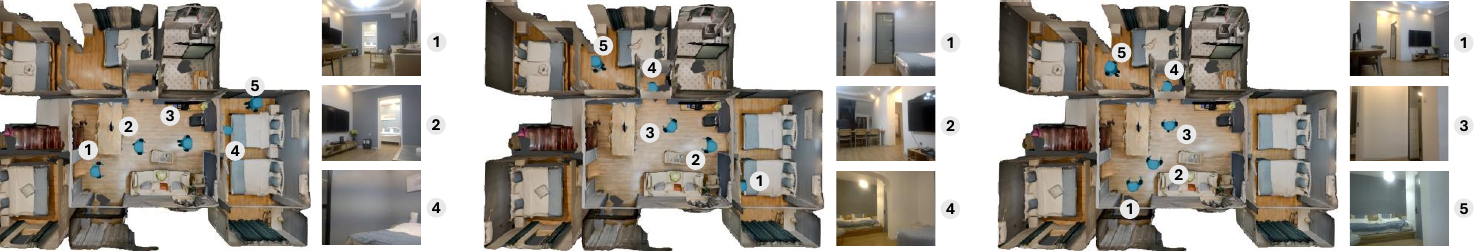}
    \caption{\textbf{Three real-world test episodes.} The left columns present the top-down view of successful episodes, with $1$ and $5$ indicating the start and goal positions, respectively. The right columns show the AGV's first-person RGB observations at the corresponding positions.}
    \label{fig:realworld}
\end{figure*}

\textbf{Quantitative results.} As outlined in \cref{tab:ablation}, \model (GT) achieves the best performance across all cases. Two key conclusions can be drawn. 
First, the removal of either the local guidance or the global guidance significantly degrades the model's performance. This indicates that both global and local guidance contribute to enhancing the robot's ability to navigate safely in complex environments. Second, w/o depth model outperforms w/o path model across SR and SPL metrics. This suggests that explicit global planning based on the floor plan provides more valuable information for efficient navigation than the local depth information derived from observations.
\begin{table}[t!]
    \centering
    \caption{Performance Comparison of w/o Perturbation and \model}
    \label{tab:robustness of gt model}
    \resizebox{\linewidth}{!}{%
    \begin{tabular}{ccccccc}%
        \toprule
        \multicolumn{3}{c}{\multirow{2}[4]{*}{Model}} & \multicolumn{2}{c}{w/o perturbation} & \multicolumn{2}{c}{\model} \\
        \cmidrule{4-7}
        & & \multicolumn{1}{c}{} & SR & SPL & SR & SPL \\
        \midrule
        \multicolumn{1}{c|}{\multirow{4}{*}{$u_1 (m)$}} & \multicolumn{1}{c}{\multirow{4}{*}{$\tau_c$}}     
                                                        & $10$     & $61.00$      & $42.08$    & $73.00$ & $53.95$ \\
        \multicolumn{1}{c|}{}  & \multicolumn{1}{c}{}   & $30$     & $73.00$      & $57.15$    & $86.00$ & $61.86$ \\
         \multicolumn{1}{c|}{}  & \multicolumn{1}{c}{}  & $50$     & $80.00$      & $61.79$    & $92.00$ & $64.84$ \\
        \multicolumn{1}{c|}{}  & \multicolumn{1}{c}{}   & $\infty$ & $86.00$      & $63.41$    & $96.00$ & $65.63$ \\     
        \midrule
        \multicolumn{1}{c|}{\multirow{4}{*}{$u_2 (m)$}} & \multicolumn{1}{c}{\multirow{4}{*}{$\tau_c$}}
                                                        & $10$     & $42.00$      & $29.90$    & $65.00$ & $44.40$ \\
        \multicolumn{1}{c|}{}  & \multicolumn{1}{c}{}   & $30$     & $52.00$      & $34.41$    & $69.00$ & $52.07$ \\
        \multicolumn{1}{c|}{}  & \multicolumn{1}{c}{}   & $50$     & $60.00$      & $37.73$    & $77.00$ & $57.23$ \\
        \multicolumn{1}{c|}{}  & \multicolumn{1}{c}{}   & $\infty$ & $77.00$      & $41.61$    & $86.00$ & $61.14$ \\  
        \bottomrule
    \end{tabular}%
    }%
\end{table}

\textbf{Qualitative results.} We select two episodes, each from an unseen scene, and visualize the results in \cref{fig:ablation}. When the agent navigates to the same goal, different models generate distinct actions due to their varying capabilities. The w/o depth model encounters collisions with walls and furniture in both cases, indicating the role of local guidance in generating collision-free navigation actions. In \textit{Mobridge}, the w/o path model generates incorrect actions at the intersection due to the lack of guidance from the optimal path. 

\paragraph{Ablation Study on GT Perturbation}
We retrain our model removing the GT perturbation trick mentioned in \cref{training} and refer to this revised version as w/o perturbation. To evaluate the robustness of \model and w/o perturbation to localization noise, we conduct a set of experiments across ten unseen environments. We introduce two levels of noise, $u_1 (m)\sim\mathcal{U}(0.0,0.5)$ and $u_2 (m)\sim\mathcal{U}(0.0,1.0)$, to the GT localization results and evaluate the performance of \model and w/o perturbation under both conditions. As shown in \cref{tab:robustness of gt model}, under noisy ground-truth localization, \model consistently outperforms w/o perturbation in most scenarios. This indicates that incorporating noise during training significantly enhances the model’s robustness to localization errors.

\paragraph{Ablation Study on Prediction Horizon}
We examine how different action prediction horizons affect the performance of \model (GT). Specifically, we evaluate three configurations with prediction lengths of 16, 32, and 48 steps. All models are trained for 20 epochs until convergence. We evaluate all three models using a fixed distance threshold $\tau_d = 0.25$, and vary the intervention count threshold $\tau_c$ among {10, 30, 50, $\infty$}. According to \cref{tab:steps}, all three configurations yield comparable performance, with the 32-step model performing slightly better. 

\begin{table}[t!]
    \centering
    \small
    \caption{Performance Comparison of Different Prediction Horizons}
    \label{tab:steps}
    \resizebox{\linewidth}{!}{%
    \begin{tabular}{cccccccc}%
        \toprule
        \multicolumn{2}{c}{\multirow{1}[4]{*}{Steps}} & \multicolumn{2}{c}{16} & \multicolumn{2}{c}{32} & \multicolumn{2}{c}{48} \\
        \cmidrule(rr){3-8}
         & & SR & SPL & SR & SPL & SR & SPL \\
        \midrule
        \multicolumn{1}{c}{\multirow{4}{*}{$\tau_c$}}  & $10$     & $77.40$ & $62.69$ &  $79.60$ & $66.60$ & $79.00$ & $65.00$ \\
        \multicolumn{1}{c}{}                           & $30$     & $89.80$ & $67.83$ &  $92.80$ & $70.38$ & $91.80$ & $69.09$ \\
        \multicolumn{1}{c}{}                           & $50$     & $94.00$ & $68.62$ &  $94.20$ & $70.98$ & $94.40$ & $70.32$ \\     
        \multicolumn{1}{c}{}                           & $\infty$ & $97.80$ & $71.89$ &  $97.60$ & $72.36$ & $97.40$ & $72.23$ \\
        \bottomrule
    \end{tabular}%
    }%
\end{table}

\subsection{Real World Deployment}
This section presents the results of deploying the \model on an AGV in a real apartment without any fine-tuning. The AGV is equipped with a mobile base, an RGB camera, and an NVIDIA Jetson AGX Orin for model execution. To ensure stable performance, we utilize the AGV's single-line LiDAR-based odometry for pose estimation, with the robot's initial pose pre-calibrated. On the Orin device, the model achieves an inference speed of approximately $1.21\text{Hz}$. The actions generated by \model are converted into linear and angular velocities via a PD controller.

We evaluate the system over $10$ episodes with $\tau_c=0$ and $\tau_d = 20 \text{cm}$. Each pair is tested three times to ensure reliability, yielding a final success rate of $40\%$. \cref{fig:realworld} presents three navigation episodes, demonstrating the generalization of obstacle avoidance and planning to unseen real-world environments. Please refer to the accompanying demo for a more intuitive illustration.

\clearpage
\section{Conclusion}
In summary, we propose an efficient and practical visual navigation framework based on floor plans, leveraging local and global condition extraction alongside a diffusion policy. Our method addresses the challenge of efficiently integrating visual information with floor plan for optimal path planning. Meanwhile, visual odometry ensures stable localization, enhancing real-world applicability. Looking forward, we foresee the integration of multi-modal data, such as language or object information, further improving adaptability and robustness, paving the way for intelligent indoor service robots in complex environments.

{
\bibliographystyle{IEEEtran}
\bibliography{reference}
}

\end{document}